\documentclass[runningheads]{llncs}

\usepackage{eccv}

\usepackage{eccvabbrv}

\usepackage{graphicx}
\usepackage{booktabs}

\usepackage[accsupp]{axessibility}  

\usepackage[pagebackref=true,breaklinks=true,colorlinks,bookmarks=false, citecolor=Green]{hyperref}
\definecolor{mycyan}{RGB}{212, 239, 251}
\usepackage{orcidlink}

\usepackage{amsmath,amsfonts,bm}

\def\eqref#1{equation~\ref{#1}}

\def\1{\bm{1}}

\def\mH{{\bm{H}}}
\def\mI{{\bm{I}}}

\def\mS{{\bm{S}}}
\def\mT{{\bm{T}}}

\def\mV{{\bm{V}}}
\def\mW{{\bm{W}}}

\def\mZ{{\bm{Z}}}

\DeclareMathAlphabet{\mathsfit}{\encodingdefault}{\sfdefault}{m}{sl}
\SetMathAlphabet{\mathsfit}{bold}{\encodingdefault}{\sfdefault}{bx}{n}

\usepackage{colortbl}
\usepackage{xcolor}
\definecolor{mygray}{gray}{.9}
\definecolor{goldenrod}{RGB}{245,245,220}
\newlength\savewidth\newcommand\shline{\noalign{\global\savewidth\arrayrulewidth\global\arrayrulewidth 1pt}\hline\noalign{\global\arrayrulewidth\savewidth}}
\newcolumntype{a}{>{\columncolor{mygray}}c}
\usepackage{fontawesome5}
\usepackage{booktabs}
\usepackage{makecell}
\usepackage{fontawesome5}
\usepackage{lipsum}
\usepackage{comment}
\usepackage{multirow}
\usepackage{wrapfig}
\usepackage{algorithm}
\usepackage{algorithmic}
\usepackage{xfrac}  

\usepackage{graphicx}
\usepackage{color}

\definecolor{darkgreen}{rgb}{0,0.7,0}

\definecolor{mygraytext}{gray}{.75}

\def\eg{\emph{e.g.}}

\newcommand{\red}[1]{{\color{red}#1}}

\definecolor{reconcolor}{HTML}{412F8A}
\definecolor{runpei-orange}{HTML}{F35F27}
\definecolor{runpei_blue}{HTML}{14294B}
\definecolor{datacolor}{HTML}{0009BF}
\definecolor{vitcolor}{HTML}{fc8e62}
\definecolor{xycolor}{HTML}{EF98AA}
\definecolor{lightpink}{HTML}{F9F0EE}
\definecolor{deeppink}{HTML}{EBD9E4}
\definecolor{lightpurple}{RGB}{123,107,143}
\definecolor{lightyellow}{RGB}{200,180,120}
\definecolor{MorandiLightBlue}{RGB}{150,180,220}
\definecolor{MorandiLighterBlue}{RGB}{230,240,250}
\definecolor{MorandiPink}{RGB}{200, 160, 180}
\definecolor{MorandiLightPink}{RGB}{250, 240, 245}

\newcommand{\gain}[1]{\textcolor{ForestGreen}{\scriptsize$_{+\!#1}$}}
\newcommand{\loss}[1]{\textcolor{BrickRed}{\scriptsize$_{-\!#1}$}}
\newcommand{\github}{\raisebox{-1.5pt}{\includegraphics[height=1.05em]{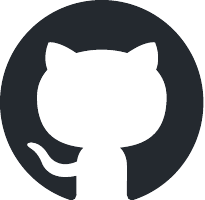}}\xspace}

\begin{document}

\title{AutoV: Loss-Oriented Ranking for Visual \\ Prompt Retrieval in LVLMs} 

\titlerunning{Loss-Oriented Ranking for Visual Prompt Retrieval in LVLMs}

\author{Yuan Zhang\inst{1,2}\orcidlink{0000-0002-1268-3131} \and
Chun-Kai Fan\inst{1}\orcidlink{0000-0002-9288-4549} \and
Sicheng Yu\inst{2}\orcidlink{0009-0008-8729-0229} \and
Junwen Pan\inst{2}\orcidlink{0000-0002-8781-6090} \and
Tao Huang\inst{3}\orcidlink{0000-0002-4463-4078} \and
Ming Lu\inst{1}\orcidlink{0000-0001-6819-6490} \and
Kuan Cheng\inst{1}\orcidlink{0000-0002-8972-1749} \and
Qi She\inst{2}\orcidlink{0000-0002-4490-2941}\and
Shanghang Zhang\inst{1}$^\dagger$\orcidlink{0000-0003-4047-3526}}

\authorrunning{Zhang et al.}

\institute{
State Key Laboratory of Multimedia Information Processing, \\School of Computer Science, Peking University \and
ByteDance \and
Shanghai Jiao Tong University
}

\begingroup
\renewcommand{\thefootnote}{}
\footnotetext{$^\dagger$ Corresponding author.}
\endgroup

\maketitle

\begin{center}
    \renewcommand{\arraystretch}{1.2}
    \begin{tabular}{rll}
        \github & \textbf{Code} & \url{https://github.com/Gumpest/AutoV}\\
    \end{tabular}
\end{center}

\begin{abstract}

Inspired by text prompts in large language models, visual prompts have been explored to enhance the perceptual capabilities of large vision-language models (LVLMs). However, performance tends to saturate under single visual prompt designs, making further prompt engineering increasingly ineffective.
To address this limitation, we shift from prompt engineering to prompt retrieval and propose AutoV, a lightweight framework for instance-adaptive visual prompt identification. Given an input image and a textual query, AutoV automatically locates the most suitable visual prompt from a diverse candidate pool.
Training such a retrieval framework requires prompt-level supervision, yet prompt quality is inherently ambiguous and difficult to assess reliably, even for humans.
To enable automatic supervision, we evaluate visual prompts using a pre-trained LVLM and label them according to their prediction losses. Using the loss-oriented ranking as a robust training signal, AutoV learns to retrieve the query-aware optimal prompt for each instance without manual annotation.
Experiments indicate that AutoV enhances the performance of various LVLMs on image understanding, captioning, grounding, and classification tasks. 
For example, AutoV improves LLaVA-OV by $\textbf{10.2}\%$ on VizWiz and boosts Qwen2.5-VL by $\textbf{3.8}\%$ on MMMU, respectively.
\keywords{Visual Prompt \and Retrieval \and Loss-Oriented \and LVLMs}
\end{abstract}

\section{Introduction}

\begin{figure*}[t]
    \centering
    \includegraphics[width=1 \textwidth]{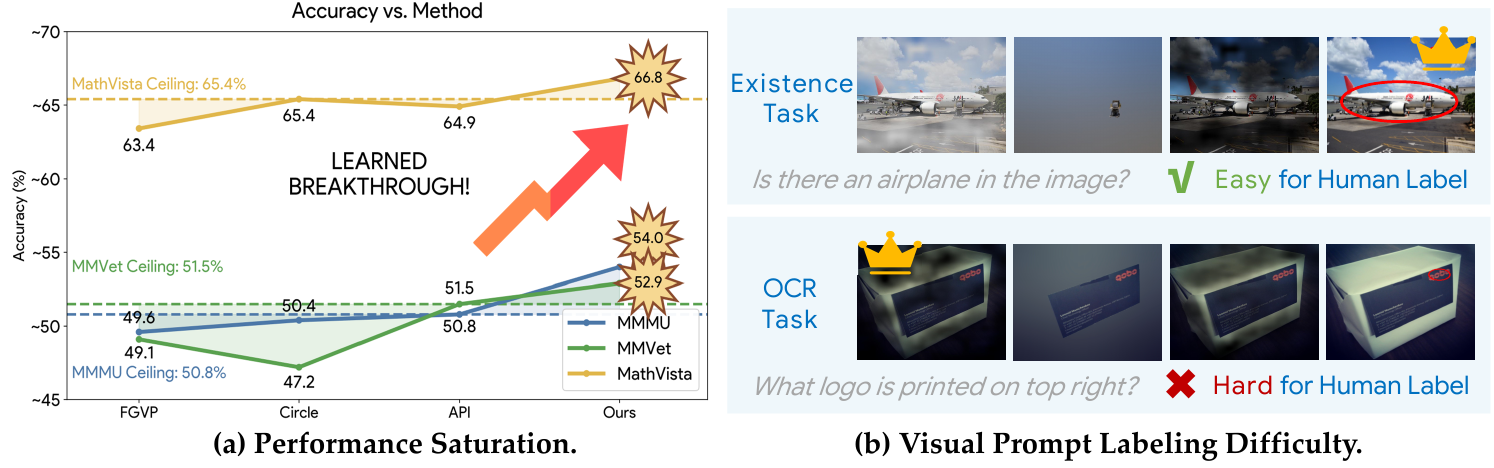}
    \vspace{-3mm}
    \caption{\textbf{Motivation of AutoV.} (a) Performance saturation. Existing visual prompts approach benchmark ceilings, limiting further gains from prompt engineering. (b) Labeling difficulty and task diversity. Optimal prompts vary across tasks, and the crown denotes the one leading to the correct answer. While the optimal prompt is easy to identify in the top example, it is much harder to determine in the bottom one.}
    \vspace{-4mm}
    \label{fig:moti}
\end{figure*}

Recent advancements in large language models (LLMs)~\cite{brown2020language, achiam2023gpt, bai2023qwen, bai2025qwen2, touvron2023llama, peng2023instruction, bi2024deepseek} have spurred significant progress in large vision-language models (LVLMs), which connect visual encoders with language model decoders to effectively adapt textual reasoning capabilities to visual understanding tasks~\cite{liu2023improvedllava, Qwen-VL, chen2023internvl, hurst2024gpt, li2024llava, team2024gemini, team2023gemini}. Consequently, visual input plays a crucial role in LVLMs, enabling precise perception and fine-grained grounding across various multimodal scenarios.

Inspired by textual prompting in LLMs~\cite{kojima2022large, wei2022chain, zhou2022cocoop}, visual prompting has emerged as an effective paradigm for guiding LVLM attention \cite{zhang2023vpgtrans, sun2023alphaclip, lin2024drawandunderstand, wu2024visual}. By injecting structured visual cues, such as blur masks, bounding circles, or attention heatmaps, visual prompts encourage models to focus on task-relevant regions. However, existing approaches predominantly rely on fixed, heuristically designed prompts. While these handcrafted strategies can yield improvements on certain benchmarks, they implicitly assume that a single prompt design generalizes across diverse visual scenes and textual queries.

In practice, this assumption rarely holds. As shown in Figure~\ref{fig:moti}(a), heuristic prompts tend to approach benchmark-specific performance ceilings, leaving limited room for improvement through prompt engineering alone. Moreover, prompt effectiveness varies across tasks and instances. For example, in Figure~\ref{fig:moti}(b), attention-based masks may benefit OCR-sensitive tasks, while highlight circles emphasize salient objects for detection. These observations suggest that \textbf{the optimal visual prompt is inherently instance-dependent}, making fixed prompt designs fundamentally limited. This insight motivates a paradigm shift from prompt engineering to prompt retrieval: instead of searching for a universally optimal visual prompt, we ask a more flexible question—given a specific image-query pair, which visual prompt is most suitable?

To this end, we introduce \textbf{AutoV}. This innovative visual-prompting retrieval framework dynamically selects the optimal visual prompt from a set of candidates, explicitly tailored to the textual query and the input image. Specifically, we first generate various visual prompt candidates and encode their representations. This diversity enables the framework to capture distinct visual representations that may be optimal for various query-image pairs. Next, we design a lightweight ranking network that efficiently integrates visual and textual information to predict the ranking preferences over prompt candidates. Finally, the highest-ranking candidate is chosen as the input for LVLMs during inference.

\textbf{A central challenge in learning such a retrieval framework lies in supervision.} As illustrated in Figure~\ref{fig:moti}(b), the quality of visual prompts is often ambiguous and task-dependent, making reliable human annotation difficult, particularly for OCR or reasoning-heavy tasks. To address this issue, we introduce a fully automated supervision strategy: each prompt candidate is evaluated by a pre-trained LVLM and ranked according to its prediction loss. This loss-oriented ranking provides a stable training signal, enabling the ranking network to learn query-aware prompt preferences without manual annotation.

Extensive experiments demonstrate that AutoV consistently enhances a wide range of LVLMs across diverse tasks, including image understanding, captioning, grounding, and classification. For example, AutoV improves LLaVA-OneVision by $\textbf{10.2}\%$ on VizWiz and boosts Qwen2.5-VL by $\textbf{3.8}\%$ on MMMU. Beyond benchmark gains, AutoV generalizes robustly across closed-source models and integrates seamlessly without additional fine-tuning of the backbone LVLM.

In summary, our contributions are threefold:

\begin{itemize}
\item We introduce AutoV, a novel automatic visual prompting framework for large vision-language models that adaptively retrieves the optimal visual prompt from a diverse candidate pool, explicitly tailored to textual query.
\item We create a scalable and totally automated data collection pipeline equipped with a reward-based loss to effectively train a lightweight ranking network, facilitating accurate and adaptive visual prompt selection without the need for extensive manual annotations.
\item We perform comprehensive experiments to validate the efficacy and robustness of AutoV across tasks, with consistent and significant improvements over existing visual prompt engineering. Once trained, AutoV integrates seamlessly into various LVLMs without additional fine-tuning.
\end{itemize}

\section{Related Work}

\subsection{Large Vision-Language Models} Recent advances in large language models \cite{radford2019language, brown2020language, achiam2023gpt, touvron2023llama, peng2023instruction, bi2024deepseek}  have catalyzed a new wave of large vision-language models \cite{liu2023improvedllava, alayrac2022flamingo, Qwen-VL, chen2023internvl, li2024mini, li2023blip}. By coupling a high-capacity visual encoder with an LLM decoder, LVLMs transfer textual reasoning to visual understanding, and the quality of input images determines the lower bound of multimodal abilities. For instance, the popular LLaVA-OneVision \cite{li2024llava} and Qwen2.5-VL \cite{bai2025qwen2} adopt a higher and dynamic resolution recipe for training to enhance their fine-grained reasoning, while vision-centric scaling in InternVL2 \cite{cai2024internlm2} enlarges the ViT \cite{dosovitskiy2020image} backbone to boost grounding capability. However, the above methods require training times of several hundred GPU days. Here, \textbf{can we further boost the existing LVLMs by optimizing the visual input itself during the inference stage?} The answer is \textbf{Yes}. Visual prompting \cite{sun2023alphaclip, lin2024drawandunderstand, jiang2024joint, zhang2024exploring, api, yang2023fine, shtedritski2023does} is an emerging and effective technique in this direction.

\subsection{Visual Prompting for LVLMs} LVLMs relying solely on textual inputs exhibited two key limitations: visual hallucinations \cite{bai2024hallucination, huang2024visual} and language-driven biases \cite{wu2024commit, qu2024unified}.
To better mitigate these challenges, visual prompting \cite{zhang2023vpgtrans, sun2023alphaclip, lin2024drawandunderstand, wu2024visual} has emerged as an alternative way of supplying cues in the visual modality. Compared with textual prompts, it is more concise and explicit, while allowing guidance that reaches fine-grained, pixel-level details. 
Early work \cite{shtedritski2023does} shows that simply drawing a red circle around a target object steers CLIP \cite{CLIP} attention, yielding great performance on referring-expression and keypoint-localization benchmarks. FGVP \cite{yang2023fine} extends this idea with fine-grained prompting, replacing coarse shapes with blur-reversal prompts based on segmentation masks, further improving accuracy on RefCOCO \cite{zhang2019referring}. AlphaCLIP \cite{sun2023alphaclip} trains an augmented version of CLIP with an auxiliary alpha channel for indicating regions at the expense of increased training overhead. API \cite{api} proposes attention prompting on images, overlaying text-guided attention heatmaps. Although previous studies have explored various visual prompts, the marginal gains brought by such engineering designs are gradually diminishing. 
To address this limitation, we propose a visual prompt retrieval mechanism that integrates existing prompt designs, exploiting their complementary advantages.

\subsection{Visual Retrieval for LVLMs} Existing visual prompt retrieval pipelines generally adhere to three main paradigms. (1) The heuristic paradigm identifies benchmark-level optimal parameters through extensive evaluation without instance-specific adaptation. (2) The random paradigm offers a baseline characterized by instance-level variation. (3) The Mixture-of-Experts (MoE) paradigm \cite{fedus2022switch, bao2022vlmo, zong2024mova} provides the most direct form of adaptation by utilizing GateNet specifically for visual retrieval.
Notably, AutoV differs fundamentally from retrieval-augmented methods~\cite{qi2024rora, xing2025re}. 
Instead of retrieving external visual data with additional finetuning, AutoV performs lightweight in-model prompt selection based solely on the input image features within a compact candidate pool, resulting in significantly improved efficiency.

\section{Proposed Approach: AutoV}
\label{sec:method}

\begin{figure*}[t]
    \centering
    \includegraphics[width=1.01\textwidth]{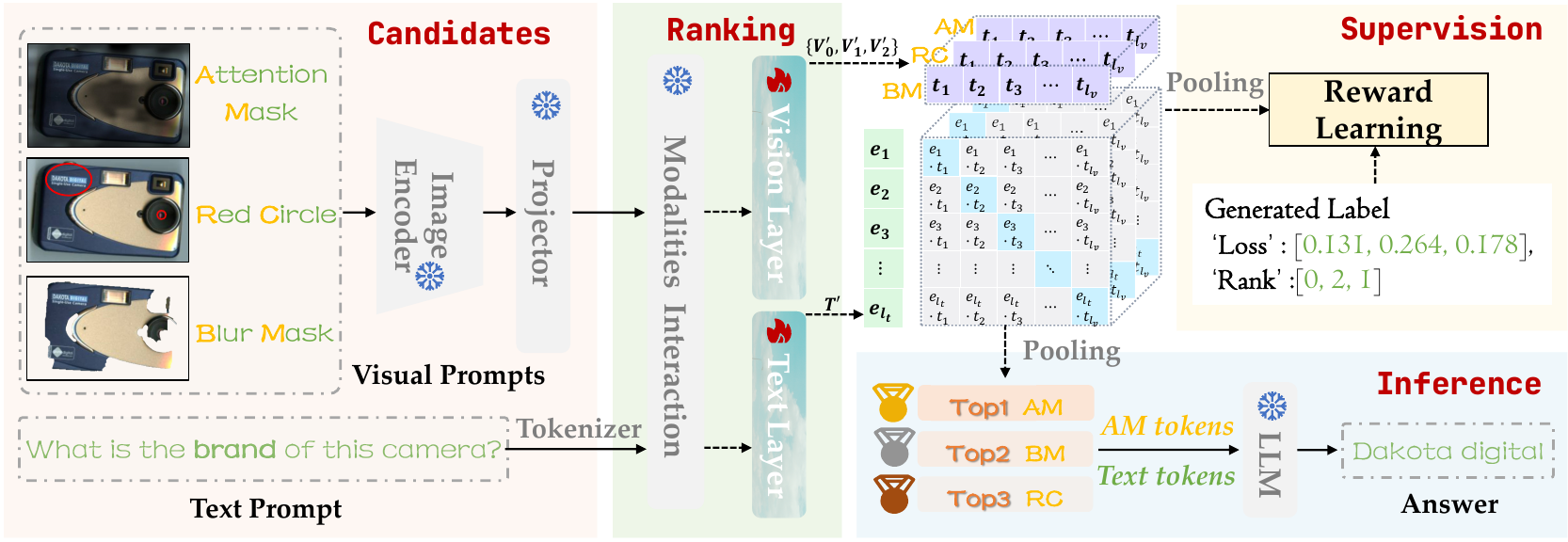}
    \vspace{-4mm}
    \caption{\textbf{The illustration of AutoV.} It comprises four key components: representation extraction from candidates, ranking network for reward score, reward-supervised training, and inference. It retrieves the prompt tailored to each query-image pair.}
    \vspace{-4mm}
    \label{fig:main}
\end{figure*}

In this section, we propose the AutoV for optimal visual prompt retrieval. Our framework includes four critical components: (1) feature extraction of prompt candidates, (2) candidate ranking consisting of modality interaction and single-modality mapping modules, (3) reward loss supervision along with automated data generation pipeline, and (4) an efficient and robust inference pipeline. The overall architecture is illustrated in Figure~\ref{fig:main}, where the four modules are color-coded for clarity. Best viewed in color.

\subsection{Feature Extraction of Prompt Candidates}
The primary goal is to effectively choose the most appropriate visual prompt in response to various text queries and input images. For a group of $n$ input visual prompts $\{x_1, x_2, ..., x_n\}$, we employ a visual encoder $g(.)$ (\emph{e.g.}, like CLIP \cite{CLIP}) to generate the visual feature: 
\begin{equation}
    \mZ_i := g(x_i) \ \text{with} \ i \in\{1, ..., n\}.
\end{equation}
Next, we apply a projection matrix $\mW$ to transform each visual feature $\mZ_i$ into language embedding tokens $\mV_i = \mW \times \mZ_i$, aligning the same dimensionality as the word embedding space in the language model. As a result, we obtain a set of visual candidates represented in the form of visual tokens $\mV_i \in \mathbb{R}^{L\times D}$ to rank, where $L$ denotes the length of vision tokens and $D$ is the language embedding dimension. Notably, both the visual encoder and the alignment projector are directly inherited from a fully trained LVLM, and we adopt them without any additional fine-tuning.

\subsection{Candidate Ranking Network}
The goal of the ranking network is to effectively rank the visual prompt candidates based on their relevance to the textual query, by modeling the relationship between the visual tokens and the input text. Therefore, we first propose the modality interaction module to enable visual tokens to fuse the context information of text tokens. Inspired by observations in \cite{zhang2025llava} that modality-fusion occurs in the early layers of the LLM, we use the first layer of the LLM decoder $f(.)$ for modality interaction. Thus, visual candidate tokens $\mV_i$ and text tokens $\mT$ are concatenated and fed into the interaction module:
\begin{equation}
    \mH_i := f(\text{concat(\(\mV_i, \mT\)})),
\end{equation}
with the outputs are $\overset{\sim}{\mV}_i = \mH_i[:l_v]$ and $\overset{\sim}{\mathbf{T}} = \mH_i[-l_t:]$. Next, we input visual candidate tokens $\overset{\sim}{\mV}_i$ into a vision mapping module, which is implemented using a feed-forward network (FFN). The generated final candidate features are ${\mV}^{'}_i \in \mathbb{R}^{l_v\times h}$, where $h \ll D$ is the output dimension of the vision mapping module to save computational cost. Besides, the query tokens $\overset{\sim}{\mathbf{T}}$ are also converted by a text mapping module to get the query embeddings $\mT^{'} \in \mathbb{R}^{l_t\times h}$. In AutoV, we define visual retrieval as a customized ranking task rather than a classification or regression problem, and the reason is discussed in Section \ref{train}. Therefore, we calculate the reward scalar for given visual candidate tokens ${\mV}^{'}_i$, where the context similarity with the query serves as the quantitative target. 
We utilize a cross-attention operation to achieve:
\begin{equation}
    s(\text{VP}_i) := \text{mean}(\text{cross-attn}({\mV}^{'}_i, \mT^{'})),
\end{equation}
where $\text{VP}_i$ denotes the $i$th visual prompt candidate. Here, only the vision and text mapping modules need additional training, each consisting of a single FFN with two linear layers. The modules comprise $2h(D + h + 2)$ parameters in total, thereby resulting in lower training costs.

\begin{figure*}[t]
    \centering
    \includegraphics[width=1\textwidth]{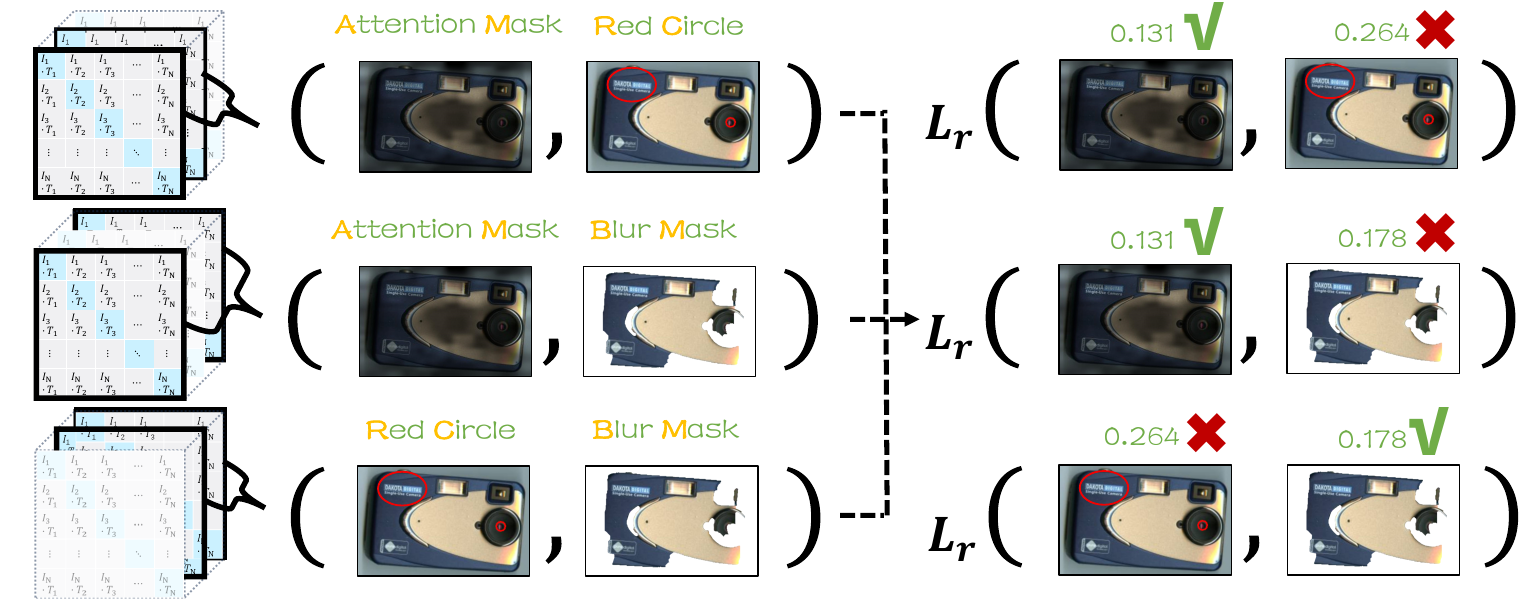}
    \vspace{-3mm}
    \caption{\textbf{The reward loss of AutoV.} As a concrete example, we illustrate the \textbf{pairwise} combination process using the visual prompts from Figure \ref{fig:main}, demonstrating how our supervision signal is applied.}
    \vspace{-4mm}
    \label{fig:loss}
\end{figure*}

\subsection{Reward Loss Supervision}
\label{train}

\textbf{Automated Data Generation.} Training a retrieval model requires prompt-level supervision, yet the quality of a visual prompt is inherently ambiguous and lacks a reliable evaluation criterion, \textbf{even for humans}. To address this, we construct supervision signals directly from the intrinsic behavior of a pre-trained LVLM. Our approach is grounded in a simple intuition: \textit{a superior visual prompt should induce lower conditional language modeling loss given the visual-question pair.} Rather than relying on one-hot correctness signals, which may reflect the LVLM’s language priors instead of true visual grounding \cite{xiao2024can, yu2024frame}, we optimize the ranking network using relative losses. In this way, the loss serves as a calibrated proxy for model confidence, where \textit{lower loss implies stronger alignment between the prompt and the input instance} \cite{guo2017calibration, zhang2024unveiling}.

For each group of $n$ visual candidates, we generate $n$ (\textit{VP}, \textit{query}) pairs and comprehensively evaluate them. Each pair is processed by a fully trained vision-language model to compute its \textit{combination loss}, defined as the modeling loss relative to the ground truth, and sorted to get a rank. To ensure high-quality ranking data, we implement two filtering criteria for (\textit{VP}, \textit{query}) pairs: (1) we remove pairs exhibiting low loss variance, as their insensitivity to visual prompts suggests answers are generated from the language priors rather than actual input content; (2) we exclude pairs with excessively high average loss, which indicates outlier case or weak visual prompt-query correlation.

For each raw (\textit{image}, \textit{query}) pair, we generate a training sample structured as a \textbf{\textit{quadruple}}:
$$ <\textit{query}, \{\textit{vp}_0, \textit{vp}_1,...,\textit{vp}_{n-1}\}, \textit{rank}, \textit{reward loss}>,$$
with an illustrative example provided in Figure \ref{fig:main}, and the dataset we adopt is detailed in Appendix \red{A}.

\noindent\textbf{Reward Loss Supervision.} Inspired by the reward modeling from human feedback in OpenAI \cite{ouyang2022training}, we train our ranking network using reward modeling to align its partial order with the optimal visual prompt. Here, we model visual retrieval as a customized ranking task rather than a classification or regression problem. This is because the objective is not to assign absolute labels or scores, but to identify the most appropriate prompt from a set of candidates by comparing their relative effectiveness in enhancing multimodal understanding. Ranking allows the model to learn fine-grained preferences across diverse query-image combinations and better align with the real-world setting. 

In Figure \ref{fig:loss}, we pair all $n$ visual prompt candidates exhaustively, ensuring each prompt is combined with every other prompt exactly once. This results in generating $\mathcal{C}(n, 2)$ training pairs for each instance. The reason is that, unlike pointwise scoring, which assigns low scores to all responses for inherently challenging tasks, pairwise comparison allows for finer discrimination between responses. Based on the rank and loss from the quadruples, we reinforce the visual prompt with lower loss (chosen sample) for each pair, while penalizing the higher-loss visual prompt (rejected sample).

Specifically, the reward loss $\mathcal{L}_{r}$ is designed as:
\begin{equation}
    \mathcal{L}_{r} :=-\frac{1}{\mathcal{C}(n, 2)} \mathbb{E} \bigg[ \log \Big( \sigma \big( s(\text{VP}_c) - s(\text{VP}_r) \big) \Big) \bigg],
    \label{eq:loss}
\end{equation}
where $s(\text{VP}_c)$ is the scalar output of the ranking network for the chosen visual prompt out of the pair, $s(\text{VP}_r)$ represents the rejected one, and $\sigma$ denotes the Sigmoid function.

\subsection{Robust Inference Pipeline}
\label{inference}
During the inference, we input multiple candidate visual prompts along with the query into the large vision-language model and pass them through our ranking network, as shown in Figure~\ref{fig:main}. The network outputs a reward scalar for each visual prompt, and we select the highest-scoring one. Its visual tokens are concatenated with the text embeddings and decoded by the LLM to generate the final answer. 
Besides, \textbf{to reduce distributional bias}, we introduce a pre-filtering step: the prompt whose visual features are farthest from other candidates, as measured by cosine distance, is removed in advance, further mitigating the negative effects of suboptimal prompts and enhancing robustness.

\section{Experiments}
\label{sec:exp}

\subsection{Experimental Setup}

\textbf{Model architectures.} We apply AutoV to four open-source LVLMs, including the LLaVA-1.5 series \cite{liu2024visual}, LLaVA-OneVision \cite{li2024llava}, Qwen2.5-VL \cite{bai2025qwen2}, and InternVL2 \cite{chen2024expanding}. 
In particular, the LLaVA-1.5 builds on Llama2 \cite{touvron2023llama2}. LLaVA-OneVision enhances it with SigLIP \cite{siglip} and Qwen2 \cite{bai2023qwen}. Qwen2.5-VL introduces a dynamic-resolution encoder, and InternVL2 integrates InternViT \cite{chen2023internvl} with InternLM2 \cite{cai2024internlm2}.

\noindent\textbf{Evaluation benchmarks.} 
AutoV is extensively evaluated on \textbf{fourteen} vision-language benchmarks, covering video reasoning (CVBench \cite{tong2024cambrian}), vision-centric tests (BLINK \cite{fu2024blink}, MMVP \cite{tong2024eyes}, MMStar \cite{chen2024we}), comprehensive multimodal evaluation (MM-Vet \cite{yu2023mm}), knowledge/math reasoning (MMMU \cite{yue2024mmmu}, MathVista \cite{lu2023mathvista}), robustness and VQA (VizWiz \cite{bigham2010vizwiz}, Real-World QA \cite{RealWorldQA}, TextVQA \cite{singh2019towards}, LLaVA-Wild \cite{liu2023improvedllava}), grounding (RefCOCO+ \cite{yu2016modeling}), captioning (COCO-Caption2017 \cite{lin2014microsoft}), and multimodal classification (ImageNet-1K \cite{deng2009imagenet}).

\renewcommand{\multirowsetup}{\centering}
\definecolor{mygray}{gray}{.95}

\definecolor{ForestGreen}{RGB}{34,139,34}
\newcommand{\fg}[1]{\mathbf{\mathcolor{ForestGreen}{#1}}}
\definecolor{Forestred}{RGB}{220,50,50}
\newcommand{\fr}[1]{\mathbf{\mathcolor{Forestred}{#1}}}

\definecolor{GreenL}{rgb}{0.56, 0.74, 0.56}
\newcommand{\fgL}[1]{\mathbf{\mathcolor{GreenL}{#1}}} 

\begin{table*}[t]
    \renewcommand{\arraystretch}{1}
    \footnotesize
	\centering
	\caption{
    \textbf{Comparison of AutoV with other visual prompts methods for various LVLMs.} \textcolor{Forestred}{Red} indicates a decrease, while \textcolor{ForestGreen}{green} represents an improvement.}
    \vspace{-2mm}
	\label{tab:main}
\resizebox{1.\linewidth}{!}{
\begin{tabular}{l | c c c c c c c}
\shline
\textbf{Method} & \textbf{MMMU} & \textbf{VizWiz} & \textbf{MMVet} & \textbf{VQA}$^{\text{Text}}$ & \textbf{LLaVA}$^{\text{Wild}}$ & \textbf{MMStar} & \textbf{MathVista} \\
\hline

\multicolumn{8}{c}{\raisebox{-0.2\height}{\includegraphics[height=1.em]{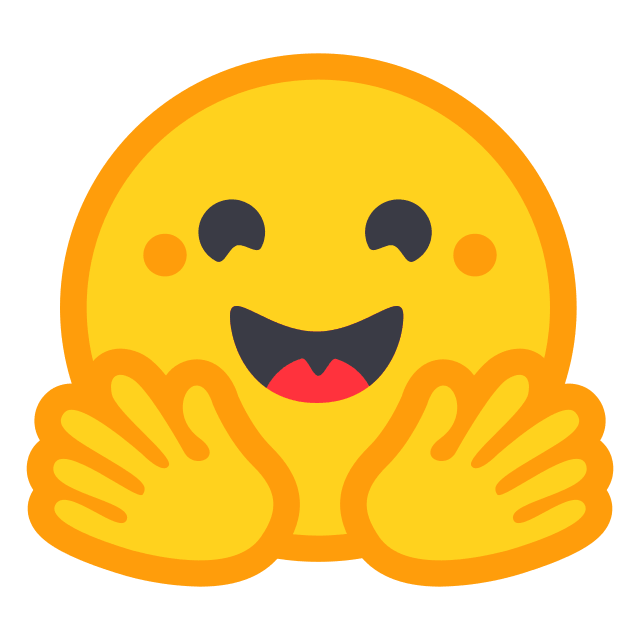}}\hspace{0.2em} \textbf{(LLaVA-1.5 7B)}}\\
Base & 36.3 & 50.0 & 30.6 & 58.2 & 65.4 & 33.2 & 34.2 \\
FGVP  & 36.9 \gain{0.6} & 47.5 \loss{2.5} & 29.0 \loss{1.6} & 46.3 \loss{11.9} & 55.6 \loss{9.8} & 33.0 \loss{0.2} & 34.6 \gain{0.4}\\
Circle  & 37.0 \gain{0.7} & 49.6 \loss{0.4} & 31.3 \gain{0.7} & 55.5 \loss{2.7} & 62.2 \loss{3.2} & 33.4 \gain{0.2} & 35.0 \gain{0.8}\\
API  & 37.4 \gain{1.1} & 51.3 \gain{1.3} & 31.8 \gain{1.2} & 58.4 \gain{0.2} & 67.2 \gain{1.8} & 34.0 \gain{0.8} & 35.7 \gain{1.5}\\
\textbf{AutoV} & \textbf{38.7} \gain{2.4} & \textbf{52.1} \gain{2.1} & \textbf{32.9} \gain{2.3} & \textbf{60.6} \gain{2.4} & \textbf{68.6} \gain{\textbf{3.2}} & \textbf{35.2} \gain{2.0} & \textbf{36.9} \gain{2.7}\\
\hline

\multicolumn{8}{c}{\raisebox{-0.2\height}{\includegraphics[height=1.em]{fig/llava.png}}\hspace{0.2em} \textbf{(LLaVA-1.5 13B)}}\\
Base & 36.7 & 53.6 & 36.1 & 61.2 & 66.6 & 32.8 & 35.0\\
FGVP  & 36.9 \gain{0.2} & 46.8 \loss{6.8} & 29.7 \loss{6.4} & 49.2 \loss{12.0} & 63.7 \loss{2.9} & 31.9 \loss{0.9} & 35.3 \gain{0.3} \\
Circle  & 36.9 \gain{0.2} & 53.8 \gain{0.2} & 31.2 \loss{4.9} & 58.1 \loss{3.1} & 66.9 \gain{0.3} & 33.5 \gain{0.7} & 36.2 \gain{1.2}\\
API  & 37.3 \gain{0.6} & 54.6 \gain{1.0} & 37.6 \gain{1.5} & 62.1 \gain{0.9} & 67.6 \gain{1.0} & 33.0 \gain{0.2} & 37.0 \gain{2.0} \\
\textbf{AutoV} & \textbf{38.9} \gain{2.2} & \textbf{56.3} \gain{2.7} & \textbf{38.0} \gain{1.9} & \textbf{62.7} \gain{1.5} & \textbf{69.5} \gain{2.9} & \textbf{34.2} \gain{1.4} & \textbf{38.3} \gain{\textbf{3.3}}\\
\hline

\multicolumn{8}{c}{\raisebox{-0.2\height}{\includegraphics[height=1.em]{fig/llava.png}}\hspace{0.2em} \textbf{(LLaVA-OneVision 7B)}}\\
Base & 49.4 & 58.2 & 48.8 & 76.1 & 80.6 & 61.8 & 63.2 \\
FGVP  & 49.6 \gain{0.2} & 58.9 \gain{0.7} & 49.1 \gain{0.3} & 65.4 \loss{10.7} & 70.8 \loss{9.8} & 58.8 \loss{3.0} & 63.4 \gain{0.2}\\
Circle  & 50.4 \gain{1.0} & 60.2 \gain{2.0} & 47.2 \gain{1.6} & 77.2 \gain{1.1} & 80.8 \gain{0.2} & 62.7 \gain{0.9} & 65.4 \gain{2.2}\\
API  & 50.8 \gain{1.4} & 66.9 \gain{\textbf{8.7}} & 51.5 \gain{2.7} & 77.7 \gain{1.6} & 81.9 \gain{1.3} & 63.0 \gain{1.2} & 64.9 \gain{1.7}\\
\textbf{AutoV} & \textbf{54.0} \gain{\textbf{4.6}} & \textbf{68.4} \gain{\textbf{10.2}} & \textbf{52.9} \gain{\textbf{4.1}} & \textbf{78.0} \gain{1.9} & \textbf{85.3} \gain{\textbf{4.7}} & \textbf{64.5} \gain{2.7} & \textbf{66.8} \gain{\textbf{3.6}} \\
\hline

\multicolumn{8}{c}{\raisebox{-0.2\height}{\includegraphics[height=1.em]{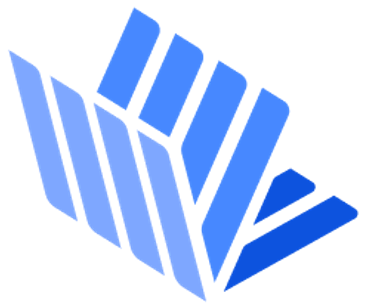}}\hspace{0.2em} \textbf{(InternVL2 8B)}}\\
Base & 48.6 & 58.6 & 46.5 & 77.0 & 74.6 & 62.0 & 58.3\\
FGVP  & 49.6 \gain{1.0} & 58.3 \loss{0.3} & 44.3 \loss{2.2} & 70.7 \loss{6.3} & 75.7 \gain{1.1} & 60.0 \loss{2.0} & 58.9 \gain{0.6}\\
Circle  & 48.3 \loss{0.3} & 57.5 \loss{1.1} & 45.8 \loss{0.7} & 74.9 \loss{2.1} & 75.4 \gain{0.8} & 63.4 \gain{1.4} & 58.4 \gain{0.1} \\
API  & 50.5 \gain{1.9} & 61.2 \gain{2.6} & 48.6 \gain{2.1} & 77.0 \gain{0.0} & 76.2 \gain{1.6} & 63.9 \gain{1.9} & 59.4 \gain{1.1}\\
\textbf{AutoV} & \textbf{51.7} \gain{\textbf{3.1}} & \textbf{64.8} \gain{\textbf{6.2}} & \textbf{50.2} \gain{\textbf{3.7}} & \textbf{78.0} \gain{1.0} & \textbf{79.5} \gain{\textbf{4.9}} & \textbf{65.8} \gain{\textbf{3.8}} & \textbf{70.5} \gain{2.2}\\
\hline

\multicolumn{8}{c}{\raisebox{-0.2\height}{\includegraphics[height=1.3em]{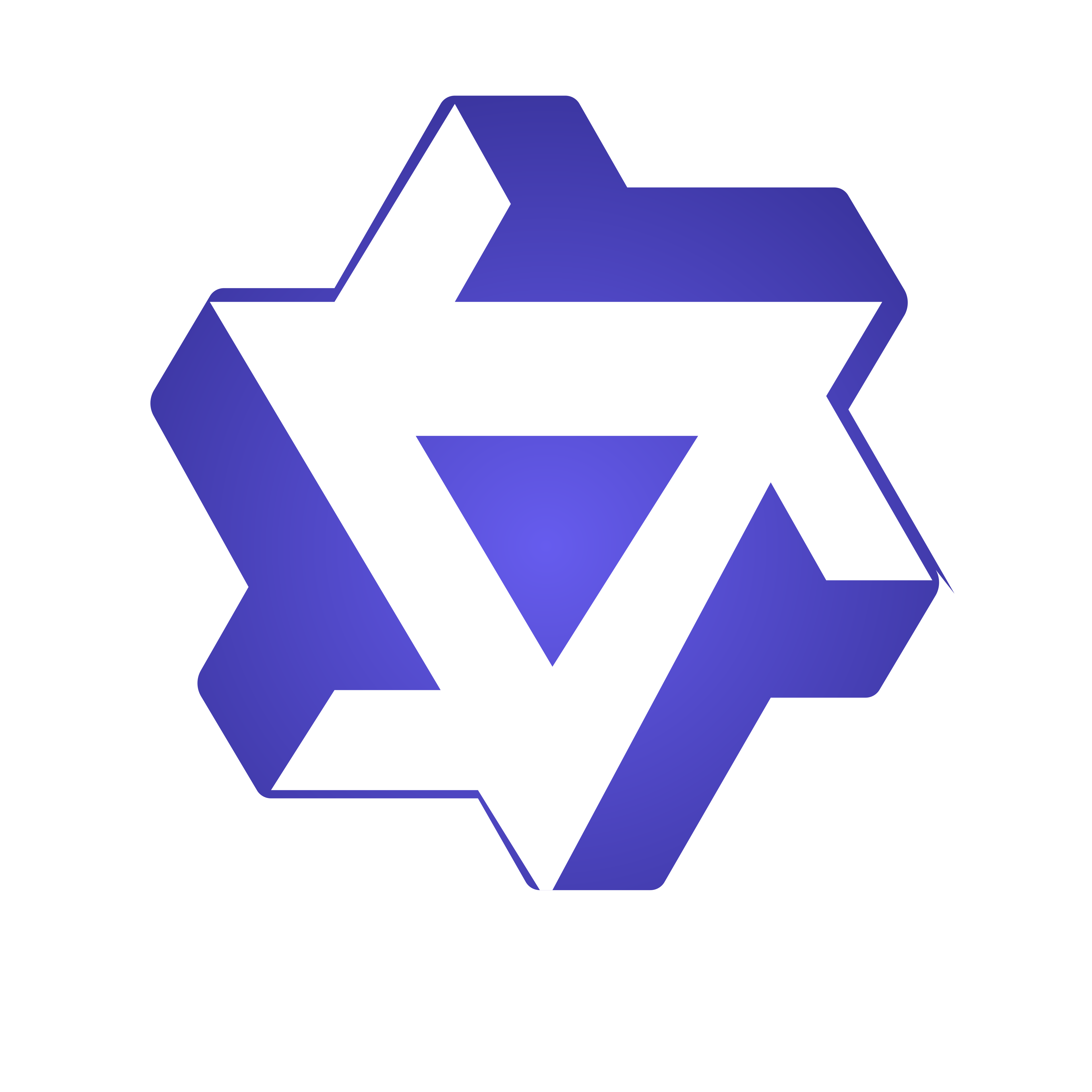}}\hspace{0.2em} \textbf{(Qwen2.5-VL 7B)}}\\
Base & 50.1 & 69.5 & 56.6 & 83.0 & 98.0 & 62.4 & 68.2\\
FGVP  & 49.6 \loss{0.5} & 70.8 \gain{1.3} & 43.1 \loss{13.5} & 78.0 \loss{5.0} & 98.2 \gain{0.2} & 63.4 \gain{1.0} & 68.0 \loss{0.2}\\
Circle  & 51.6 \gain{1.5} & 69.2 \loss{0.3} & 44.5 \loss{12.1} & 80.5 \loss{2.5} & 97.5 \loss{0.5} & 65.1 \gain{2.7} & 69.1 \gain{0.9}\\
API  & 51.0 \gain{0.9} & 72.1 \gain{2.6} & 58.0 \gain{1.4} & 85.3 \gain{2.3} & 100.6 \gain{2.6} & 63.0 \gain{0.6} & 67.6 \loss{0.6}\\
\textbf{AutoV} & \textbf{53.9} \gain{\textbf{3.8}} & \textbf{74.4} \gain{\textbf{4.9}} & \textbf{59.1} \gain{2.5} & \textbf{86.1} \gain{\textbf{3.1}} & \textbf{102.3} \gain{\textbf{4.3}} & \textbf{65.6} \gain{\textbf{3.2}} & \textbf{70.2} \gain{2.0}\\
\shline
\end{tabular}
}
\vspace{-4mm}
\end{table*}

\subsection{Main Results}

Results are presented in Table \ref{tab:main}, and we compare AutoV with three prior visual prompting methods for LVLMs, namely FGVP \cite{yang2023fine}, RedCircle \cite{shtedritski2023does}, and API \cite{api}.
The training data and implementation details can be found in Appendix \red{A} and \red{C}, respectively. Here, we employ the above three types of visual prompts as the candidate pool\footnote{Includes four prompts extracted from different transformer layers of CLIP in API \cite{api}, with two additional prompts of distinct types, resulting in six images in total.} for AutoV to retrieve from optimally.

The four main observations are as follows: 
(1) For all models, the average improvement of AutoV relative to the base models is 3.5\%. Our method performs particularly well for LLaVA-OneVision on VizWiz, with an improvement of $\textbf{10.2}\%$.
(2) AutoV outperforms existing visual prompt engineering methods. For instance, on MMMU, it achieves an average gain of $\mathbf{3.2\%}$, compared to $\mathbf{1.1\%}$ for API, yielding a $\mathbf{2.1\%}$ absolute improvement. This margin highlights the importance of accurate case-level visual cue retrieval, as opposed to benchmark-level prompt design. In contrast, other visual prompting strategies that lack query-adaptive mechanisms tend to underperform. (3) Our method further supplements inherently less accurate models. For instance, LLaVA-7B with AutoV achieves higher accuracy than the 13B-scale model on \textbf{4 out of 7 tasks}. This means that AutoV can be utilized for model efficiency on certain specific tasks: \textbf{a few additional MLP layers are a better choice compared with hundreds of times larger parameter sizes.} (4) Both Qwen2.5-VL and InternVL2 directly adopt the retrieval optimal prompting strategy from AutoV, which is pre-trained on LLaVA-OneVision. This integration results in an average accuracy gain of $\textbf{4.2}\%$. This demonstrates that the retrieval-enhanced prompting strategy is \textbf{model-agnostic} and generalizes well across heterogeneous LVLM architectures, regardless of pretraining scales or vision encoders.
Additionally, we extend the evaluation to a broader range of vision-language benchmarks in Appendix~\red{E}, highlighting AutoV's strong cross-task generalization in retrieval.

\section{Analysis}

\subsection{Comparison with Other Retrieval Methods}
\begin{table}[t]
  \caption{\textbf{Analysis between AutoV and other retrieval methods.} $\text{VP}_1$-$\text{VP}_4$ are from attention prompting \cite{api} from different layers.}
  \vspace{-2mm}
  \renewcommand{\arraystretch}{1}
  \setlength{\tabcolsep}{2.7mm}
  \label{tab:ab1}
  \centering
  \footnotesize
  \begin{tabular}{l|c|c|c|c}
    \toprule
    \textbf{Method} & \textbf{MMMU} & \textbf{VizWiz} & \textbf{MMVet} & \textbf{Avg.}\\
    \midrule
    Baseline & 36.3 & 50.0 & 30.6 & 39.0 \\
    \midrule
    $\text{VP}_1$ only & 36.7 \gain{0.4} & 50.7 \gain{0.7} & 31.8 \gain{1.2} & 39.8 \gain{0.8}  \\
    $\text{VP}_2$ only & 36.9 \gain{0.6} & 51.3 \gain{1.3} & 31.4 \gain{0.8} & 39.9 \gain{0.9} \\
    $\text{VP}_3$ only & 37.4 \gain{1.1} & 50.6 \gain{0.6} & 31.7 \gain{1.1} & 39.9 \gain{0.9} \\
    $\text{VP}_4$ only & 37.2 \gain{0.9} & 51.2 \gain{1.2} & 31.2 \gain{0.6} & 39.9 \gain{0.9} \\
    \midrule
    Random & 37.1 \gain{0.8} & 50.7 \gain{0.7} & 31.3 \gain{0.7} & 39.7 \gain{0.7} \\
    \midrule
    Regression & 36.9 \gain{0.6} & 50.9 \gain{0.9} & 31.1 \gain{0.5} & 39.6 \gain{0.6} \\
    MoE & 37.6 \gain{1.3} & 50.7 \gain{0.7} & 32.0 \gain{1.4} & 40.1 \gain{1.1} \\
    \midrule
    \textbf{AutoV} List-wise ranking & 38.0 \gain{1.7} & 51.3 \gain{1.3} & 32.4 \gain{1.8} & 40.6 \gain{1.6} \\
    \textbf{AutoV} Pair-wise ranking & \textbf{38.7} \gain{2.4} & \textbf{52.1} \gain{2.1} & \textbf{32.9} \gain{2.3} & \textbf{41.3} \gain{2.3} \\
    \bottomrule
  \end{tabular}
\vspace{-10pt} 
\end{table}
To analyze why pair-wise ranking is more suitable for visual prompt retrieval, we compare it with several representative retrieval strategies. 
All experiments are conducted on LLaVA-7B with the retrieval pool utilized in the main experiments.

{\textbf{1. Heuristic Fixed Retrieval.}} This strategy selects a single predefined visual prompt based on manually designed rules. As shown in Rows 3--6 of Table~\ref{tab:ab1}, different prompts exhibit distinct performance profiles across benchmarks. For instance, $\text{VP}_3$ performs best on MMMU, while $\text{VP}_1$ is optimal on MMVet, indicating that no single prompt engineering is universally effective. 

{\textbf{2. Random Retrieval.}}
Randomly selecting yields a +0.7\% average gain over the baseline and performs on par with or slightly better than individual fixed prompts. This suggests that prompt retrieval itself provides benefits.

{\textbf{3. Regression Retrieval.}}
This strategy directly predicts the absolute loss score of each visual prompt and selects the lowest one. As shown in Table~\ref{tab:ab1}, regression improves the baseline by $0.6\%$, but performs worse than both MoE and ranking-based methods. This suggests that learning absolute loss is challenging due to small performance gaps among prompts and inherent noise in supervision signals, making it difficult to reliably distinguish the optimal prompt.

{\textbf{4. MoE (GateNet) Retrieval.}}
We further compare with a Mixture-of-Experts (MoE) style retrieval mechanism, where a lightweight gating network (implemented with Gumbel Softmax~\cite{fedus2022switch}) learns to select among multiple visual prompts in a differentiable manner. GateNet is inserted after the projection layer and trained under the same data and training budget as AutoV. As shown in Row 9 of Table~\ref{tab:ab1}, MoE improves the baseline by $1.1\%$. While this demonstrates the benefit of learnable selection, its gain over random retrieval is modest (+0.4\%), and it remains $1.2\%$ behind AutoV (pair-wise ranking). 

{\textbf{5. List-wise Ranking.}}
We further adopt a list-wise ranking objective that models the relative ordering of all candidate prompts simultaneously. List-wise ranking achieves consistent gains (+1.6\%), demonstrating the benefit of structured preference learning over regression or soft gating. However, compared to pair-wise ranking, its improvement is more limited. We conjecture that list-wise optimization requires estimating a full permutation over candidates, 
which introduces higher optimization complexity and sensitivity to noisy comparisons.

{\textbf{6. AutoV.}}
Overall, these comparisons suggest that effective visual prompt retrieval requires 
(i) query-aware adaptability beyond random diversity, 
(ii) explicit relative preference modeling beyond soft gating, and 
(iii) structured supervision beyond list-wise ranking. 
Pair-wise ranking naturally satisfies these properties, 
which explains its consistent superiority in Table~\ref{tab:ab1}.

\begin{wrapfigure}{r}{0.40\textwidth}
    \vspace{-15mm}
    \begin{center}
        \includegraphics[width=\linewidth]{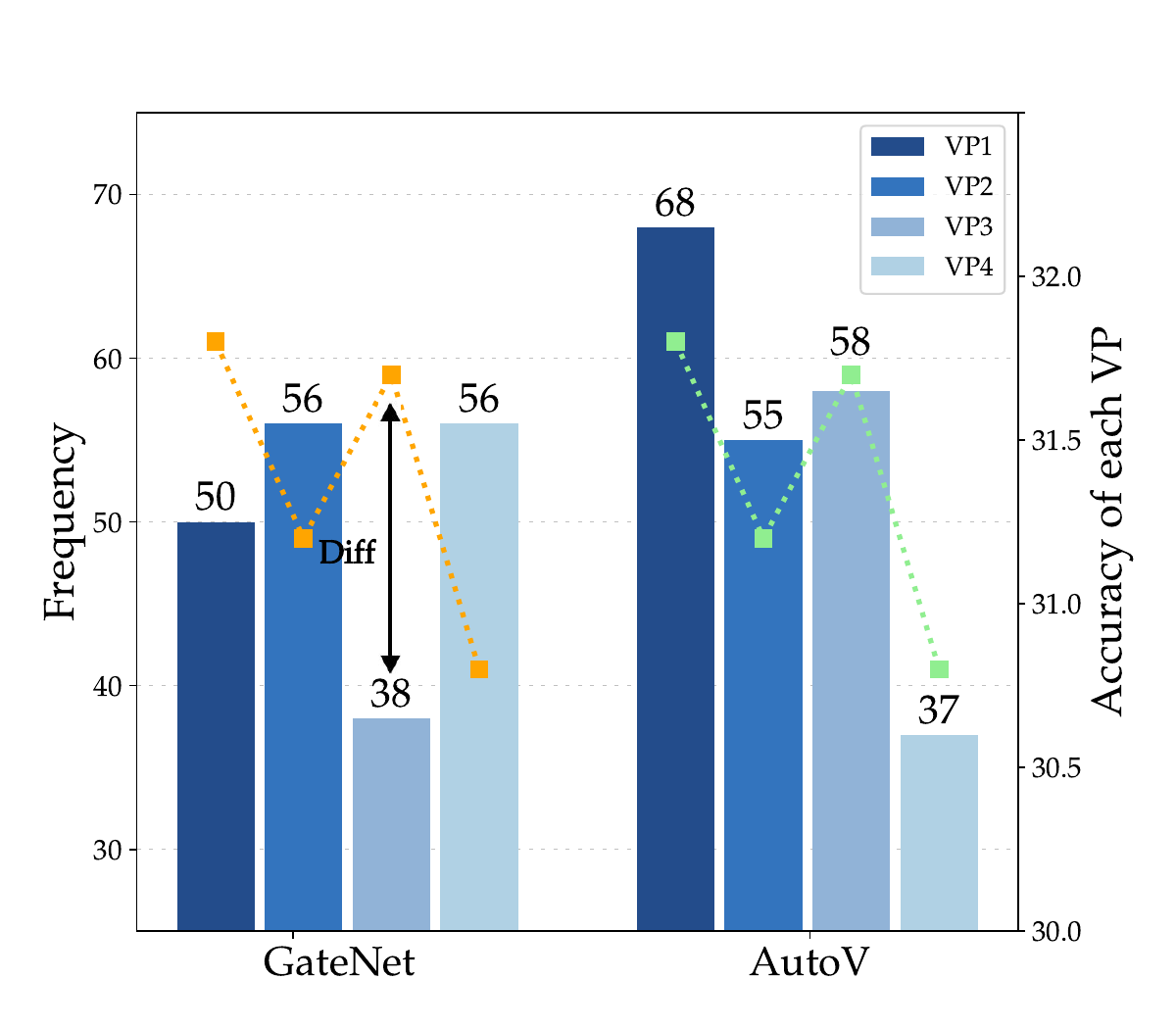}
    \end{center}
    \vspace{-4mm}
    \caption{\textbf{Retrieval distribution on MMVet.} The plots are independent accuracy for each prompt.}
    \label{fig:distribution}
    \vspace{-8mm}
\end{wrapfigure}

\subsection{Retrieval Visualization} 
We visualize the prompt distribution of retrieval strategies within $\text{VP}_1$-$\text{VP}_4$ for significance. Figure \ref{fig:distribution} presents a bar graph comparing the retrieval decisions of GateNet and our AutoV on MMVet. The bar heights indicate the frequency of visual prompt retrieval, while the overlaid line plots depict the independent accuracy for each prompt.
Our analysis reveals a notable discrepancy between the GateNet retrieval and the actual inference accuracy of visual prompts. For instance, GateNet retrieves $\text{VP}_3$, the second-most useful prompt, only $38$ times (38/218, $13.6\%$ of cases), despite its high accuracy. In contrast, AutoV exhibits stronger alignment with the standalone accuracy of visual prompts, and prefers to select $\text{VP}_1$ and $\text{VP}_3$, the two prompts that own superior accuracy.

\subsection{The Scaling Law of AutoV}

\begin{table*}[t]
  \caption{\textbf{Performance of various sizes of prompts pool.} 
  1–4 are attention prompting images \cite{api} from different layers; 
  5 and 6 come from \cite{shtedritski2023does} and \cite{yang2023fine}; 
  7 and 8 are \textbf{newly unseen prompts} (\cite{jiang2024joint} and \cite{zhang2024exploring}) 
  for analyzing scaling law effects.}
  \vspace{-1mm}
  \renewcommand{\arraystretch}{1.2}
  \setlength{\tabcolsep}{0.5mm}
  \label{tab:ab2}
  \centering
  \footnotesize
\resizebox{1.02\linewidth}{!}{
    \begin{tabular}{l|cccccc|cc}
    \toprule
    \textbf{Pool Size} & \textbf{1} & \textbf{2} & \textbf{3} & \textbf{4} & \textbf{5} & \textbf{6} & \textbf{7} & \textbf{8}\\
    \midrule
    \textbf{Source} 
    & $+$API-L$_{15}$
    & $+$API-L$_{20}$
    & $+$API-L$_{22}$
    & $+$API-L$_{23}$
    & $+$RedCircle
    & $+$FGVP
    & $+$VTP
    & $+$TVP\\
    \midrule
    
    \textbf{MMMU} 
    & 37.3\gain{1.0}
    & 37.5\gain{1.2}
    & 38.3\gain{2.0}
    & 38.3\gain{2.0}
    & 38.6\gain{2.3}
    & 38.7\gain{2.4}
    & 38.7\gain{2.4}
    & \textbf{38.9}\gain{2.6} \\
    
    \textbf{VizWiz} 
    & 50.3\gain{0.3}
    & 50.8\gain{0.8}
    & 51.0\gain{1.0}
    & 51.9\gain{1.9}
    & 52.1\gain{2.1}
    & 52.1\gain{2.1}
    & 52.6\gain{2.6}
    & \textbf{52.6}\gain{2.6} \\
    
    \textbf{MMVet} 
    & 31.8\gain{1.2}
    & 32.0\gain{1.4}
    & 32.6\gain{2.0}
    & 32.8\gain{2.2}
    & 32.9\gain{2.3}
    & 32.9\gain{2.3}
    & 33.1\gain{2.5}
    & \textbf{33.1}\gain{2.5} \\
    
    \midrule
    \textbf{Avg.} 
    & 39.8\gain{0.8}
    & 40.1\gain{1.1}
    & 40.7\gain{1.7}
    & 41.0\gain{2.0}
    & 41.2\gain{2.2}
    & 41.3\gain{2.3}
    & 41.5\gain{2.5}
    & \textbf{41.6}\gain{2.6} \\
    
    \bottomrule
    \end{tabular}
}
\vspace{-1mm}
\end{table*}
Having more visual prompt candidates increases selection diversity and potentially improves the adaptability of LVLMs. \textbf{However, is more always better?} Here, we extend the candidate pool from $1$ to $8$ prompts (Table~\ref{tab:ab2}).
\textbf{(1)} We observe a clear diminishing-return effect. When prompts are individually beneficial, increasing their number initially improves accuracy, but the marginal gain decreases as the pool grows. 
For example, on MMMU, the gain remains $0.8\%$ when increasing from $2$ to $3$ prompts, but drops to $0$ when expanding from $6$ to $7$ prompts. 
\textbf{(2)} AutoV is largely agnostic to the quality of individual visual prompts, and its effectiveness does not depend on carefully curated prompt candidates. 
On VizWiz, although RedCircle and FGVP yield negative gains ($-0.4\%$ and $-2.5\%$), incorporating them still leads to a $0.2\%$ improvement, demonstrating its robustness. 
\textbf{(3)} Moreover, even when encountering unseen prompt types during training, AutoV can effectively index suitable candidates, yielding an additional $0.3\%$ average improvement on the last two visual prompts.

Overall, AutoV remains stable under varying prompt quality and pool sizes, consistently preserving or improving performance. 
In practice, we recommend using high-quality prompts while limiting the pool to fewer than $9$ candidates to balance effectiveness and efficiency.

\subsection{Study on the LLM Layer Selection} 

\begin{table}[t]
\centering
\footnotesize
\setlength{\tabcolsep}{13pt}
\renewcommand{\arraystretch}{1.}
\caption{\textbf{Ablation studies on LLM layer selection and pre-filtering step.}}
\vspace{-2mm}
\label{tab:prefilter}
    \begin{tabular}{lccc}
    \toprule
    \textbf{Method} & \textbf{MMMU} & \textbf{VizWiz} & \textbf{MMVet} \\
    \midrule
    LLaVA-1.5-7B & 36.3 & 50.0 & 30.6 \\
    
    \rowcolor{mygray}
    \multicolumn{4}{c}{\textbf{Part1: LLM layer selection}} \\
    24th layer & 38.1 & 51.6 & 32.3 \\
    12th layer & 38.4 & 51.7 & 32.3 \\
    0th layer (\textbf{Default}) & \textbf{38.7} & \textbf{52.1} & \textbf{32.9} \\
    
    \rowcolor{mygray}
    \multicolumn{4}{c}{\textbf{Part2: Pre-filtering strategy}} \\
    
    \multicolumn{4}{c}{When 6 Candidates} \\
    \textit{w/o} pre-filtering 
        & 38.3 
        & 52.0 
        & 32.6 \\
    
    \textit{w/} pre-filtering (\textbf{Default}) 
        & \textbf{38.7}\gain{0.4} 
        & \textbf{52.1}\gain{0.1} 
        & \textbf{32.9}\gain{0.3} \\
    
    \multicolumn{4}{c}{When 8 Candidates} \\
    \textit{w/o} pre-filtering 
        & 38.4 
        & 52.2 
        & 32.7 \\
    
    \textit{w/} pre-filtering (\textbf{Default}) 
        & \textbf{38.9}\gain{0.5} 
        & \textbf{52.6}\gain{0.4} 
        & \textbf{33.1}\gain{0.4} \\
    \bottomrule
    \end{tabular}
    \vspace{-3mm}
\end{table}

Findings from FastV \cite{chen2024image} and LLaVA-Mini \cite{zhang2025llava} validate that cross-modal interactions mainly occur in shallow layers. This enables the textual input to better guide visual selection. \textbf{Part 1} of Table \ref{tab:prefilter} shows that the \texttt{0}th layer outperforms intermediate or bottom layers, validating our design choice.

\subsection{Study on the Pre-filtering Step} \textbf{Part 2} of Table~\ref{tab:prefilter} shows consistent improvements, especially with more prompts. High-recall pre-filtering removes only the most dissimilar candidates, with a negligible risk of discarding optimal prompts ($<$0.5\% in our statistics), while substantially reducing the presence of harmful prompts.

\subsection{The Transferability of AutoV}
\label{transferability}
\begin{table}[t]
    \renewcommand\arraystretch{1.}
	\setlength\tabcolsep{6mm}
    \footnotesize
    \centering
    \caption{\textbf{Performance of closed-source models with AutoV.} The retrieval strategy is generated from AutoV pre-trained on LLaVA-OneVision.}
    \label{tab:ab3}
    \vspace{-2mm}
    \begin{tabular}{l|c|c}
         \shline
         \textbf{Models} & \textbf{VizWiz} & \textbf{LLaVA}$^{\text{Wild}}$  \\
         \shline

         \raisebox{-0.25\height}{\includegraphics[height=1em]{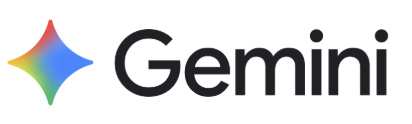}}\hspace{0.2em} (Gemini-1.5-Pro) & 45.9 & 77.9 \\
         \rowcolor{mygray}
         +AutoV & 55.5\gain{\textbf{9.6}} & 81.1\gain{\textbf{3.2}} \\
         \hline

         \raisebox{-0.25\height}{\includegraphics[height=1.05em]{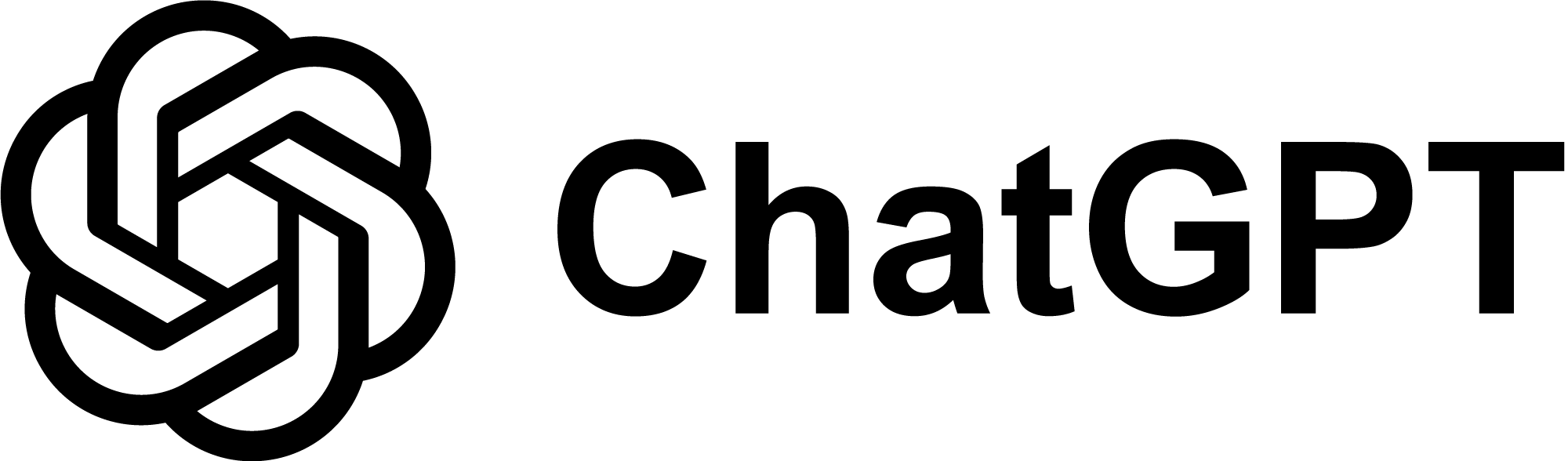}}\hspace{0.2em} (GPT-4o) & 52.8 & 68.0 \\
         \rowcolor{mygray}
         +AutoV & 61.8\gain{\textbf{9.0}} & 70.9\gain{2.9} \\
         \hline

         \raisebox{-0.25\height}{\includegraphics[height=1.05em]{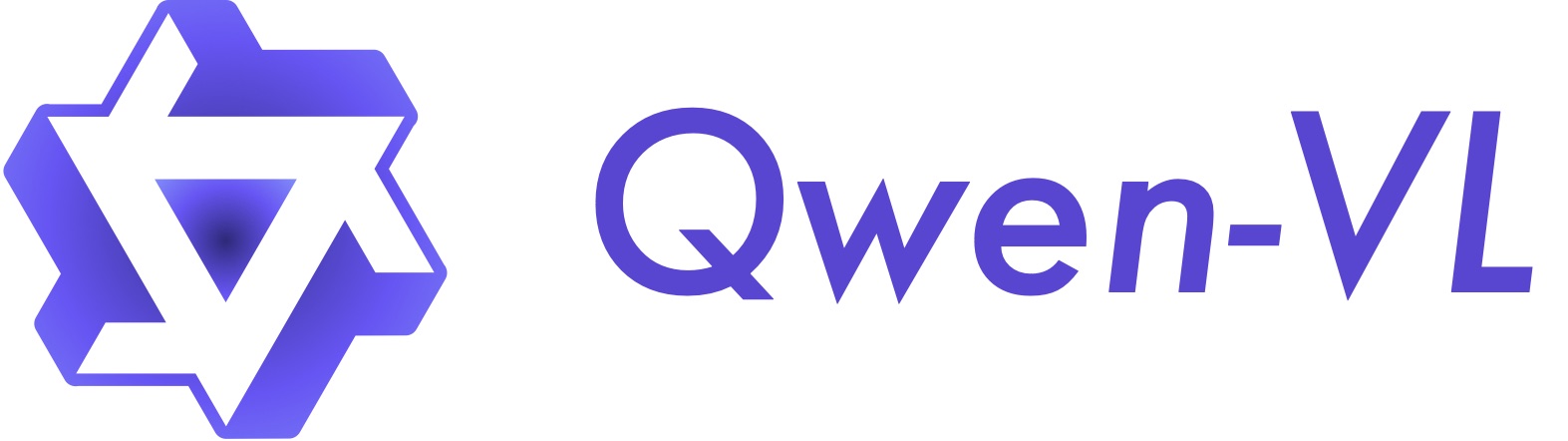}}\hspace{0.2em} (Qwen-VL-Max) & 64.6 & 85.8 \\
         \rowcolor{mygray}
         +AutoV & 71.9\gain{\textbf{7.3}} & 88.4\gain{2.6} \\
         
         \shline
    \end{tabular}
    \vspace{-2mm}
\end{table}
The retrieval strategy in our AutoV, trained on one LVLM, can be transferred to other LVLMs. First, our method demonstrates broad applicability across open-source models. In Table \ref{tab:main}, both InternVL2 and Qwen2.5-VL achieve a $3.2\%+$ improvement by directly adopting the retrieval strategy from AutoV pre-trained on LLaVA-OneVision. Besides, Table \ref{tab:ab3} reveals that even closed-source models like Gemini-Pro \cite{team2024gemini} and GPT-4o \cite{hurst2024gpt} show substantial gains, averaging $6.4$ and $6.0$ points respectively. Therefore, AutoV owns good transferability, as its retrieval strategy consistently boosts performance across both open- and closed-source LVLMs without model-specific adaptation.

\subsection{Statistical Significance Analysis}
To evaluate the reliability of our improvements, we perform paired t-tests \cite{hsu2014paired} across multiple random seeds for all benchmarks in Table~\ref{tab:main} (LLaVA-1.5 7B). Taking MMMU as an example, AutoV achieves a mean improvement of $\Delta = +2.06\%$ over the baseline with $p \leq 0.05$, indicating statistical significance rather than random fluctuation. Similar significance is consistently observed across other benchmarks. Detailed statistical results are provided in Appendix~\red{G}.

\subsection{The Efficiency of AutoV} 
\begin{table}[t]
\centering
\footnotesize
\setlength{\tabcolsep}{9pt}
\renewcommand{\arraystretch}{1.}
\caption{\textbf{Overhead analysis.} Extra and total FLOPs (T) under different pool sizes, model scales, and resolutions.}
\vspace{-2mm}
\label{tab:re_efficiency}
\begin{tabular}{lcc|cc}
\toprule
\multirow{2}{*}{\textbf{Factor}} 
& \multicolumn{2}{c|}{\textbf{Base Setting}} 
& \multicolumn{2}{c}{\textbf{Expanded Setting}} \\
& \textbf{Extra} (T) & \textbf{Total} (T)
& \textbf{Extra} (T) & \textbf{Total} (T) \\
\midrule

\textbf{Pool Size} 
& 0.74 (\textbf{4}) & 5.08 
& 1.62 (\textbf{8} $\uparrow$) & 5.96 \\

\textbf{Model Size} 
& 0.74 (\textbf{7B}) & 5.08 
& 0.74 (\textbf{13B} $\uparrow$) & 8.77 \\

\textbf{Vision Token} 
& 0.74 (\textbf{576}) & 5.08 
& 1.48 (\textbf{1152} $\uparrow$) & 6.59 \\

\bottomrule
\end{tabular}
\end{table}

\label{Efficiency}
AutoV improves LVLM performance while remaining lightweight and practical. \textbf{Training} AutoV on LLaVA-7B takes only 6 hours on 8 $\times$ A100GB GPUs (40 epochs). Thanks to its transferability, no retraining is required for new LVLMs, and training data can be generated with smaller models. \textbf{At inference}, AutoV encodes $N\!-\!1$ additional images and ranks $N$ candidates ($N \leq 8$), without extra decoder cost. For LLaVA-7B (576 visual, 20 text tokens), the encoder, ranking-net, and decoder require 0.16T, 0.09T, and 4.30T FLOPs, respectively: LLM decoding clearly dominates. Under the base setting ($N=4$), AutoV adds only 0.74T FLOPs (Table~\ref{tab:re_efficiency}). Larger pools or higher resolution increase vision-side cost nearly linearly, yet remain negligible relative to decoder computation. The overhead is independent of model scale (7B vs.\ 13B).
In practice, \textbf{latency} increases by just 6.9\,ms (254.8\,ms $\rightarrow$ 261.7\,ms) due to compact similarity computation and early prompt selection, yielding a +2.4\% average accuracy gain.

\begin{figure*}[t]
    \centering
    \includegraphics[width=1.0 \textwidth]{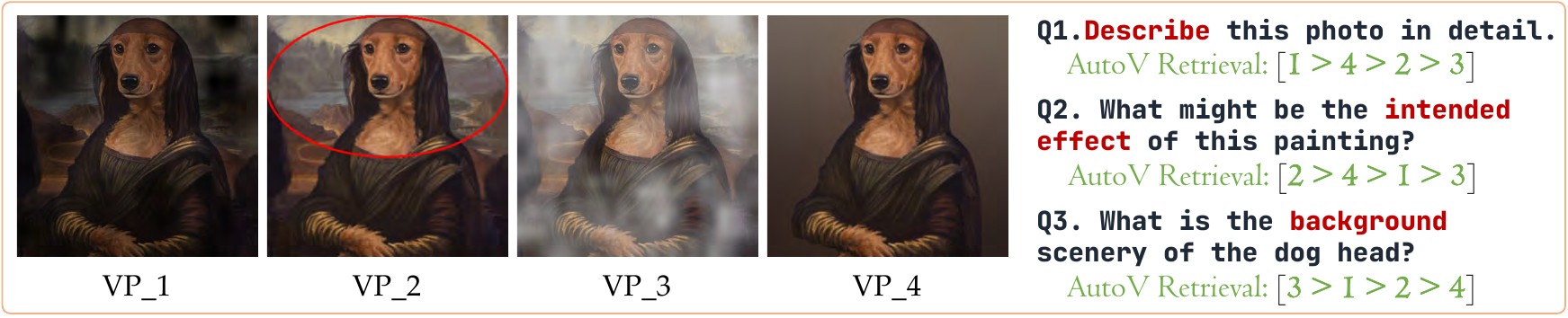}
    \vspace{-4mm}
    \caption{\textbf{Visualization of retrieval results faced with different queries.} $\text{VP}_1$ and $\text{VP}_3$ come from API, $\text{VP}_2$ is from RedCircle, and $\text{VP}_4$ is from FGVP.}
    \vspace{-2mm}
    \label{fig:vis}
\end{figure*}

\subsection{Qualitative Visualization}

To validate the effectiveness of our retrieval strategy, we visualize an example sourced from LLaVA-Bench within \{API-L$22$-Dark, RedCircle, API-L$23$-White, FGVP\}. As shown in Figure \ref{fig:vis}, we list the output results of AutoV when we enter different queries. When faced with a question that focuses on the global context information, AutoV chooses $\text{VP}_1$ (API-L$22$-Dark), where the black mask removes the background. While referring to the "intended effect" in the image, AutoV selects $\text{VP}_2$ (RedCircle), which directly uses red circles to highlight the weird spots. More visualizations can be found in Appendix \red{H}.

\section{Conclusion}

In this paper, we introduce AutoV, an adaptive framework that retrieves optimal visual prompts for textual queries in LVLMs. Leveraging a lightweight ranking network trained with reward-based supervision, AutoV consistently outperforms prompt engineering methods. Notably, its effectiveness does not rely on the intrinsic quality of individual visual prompts, nor is it constrained by their standalone performance. Extensive experiments on multiple LVLMs and benchmarks demonstrate its effectiveness, robustness, and generality.


\section*{Acknowledgements}
This work was supported by the National Natural Science Foundation of China under Grants 62476011 and 62472008, and by the Beijing Natural Science Foundation under Grant L252060.

%
%
\bibliographystyle{splncs04}
\bibliography{main}

\clearpage
\appendix
\begin{center}
    {\Large \bfseries Supplementary Material}
\end{center}

\setcounter{section}{0}
\renewcommand{\thesection}{\Alph{section}}

\section{Details of Training Data}
\label{data}

\subsection{Attention Visual Prompts for Training}
As vision foundation models advance rapidly~\cite{CLIP, zhang2025chainv, zhangsparsevlm, siglip, zhang2026freekd+, kirillov2023segment}, pretrained on language-image pairs, they serve as the visual encoders for large vision-language models. A representative model is CLIP~\cite{CLIP}, $g_{\text{CLIP}}(.)$, which consists of a vision encoder and a text encoder, mapping the image and text prompt into the aligned latent space. Therefore, we can directly obtain the similarity $\mS$ between vision modality $\mI$ and text modality feature $\mT$ with $\mS = g_{\text{CLIP}}(\mI, \mT)$. 

Inspired by the idea that text-image attention (or similarity ) $\mS$ can be visualized as visual prompts for LVLMs~\cite{api, lin2024training}, where similarity heatmaps overlaid on the image highlight key regions requiring greater attention for prompt-based reasoning. In this work, we adopt a fine-grained similarity visualization manner in API~\cite{api}, with the $l$-th layer similarity $\mS^{l}$ defined as follows:
\begin{equation}\label{eq:sim_approx}
    \mS^{l} = \text{SIM}(\mI^{l}, \mT) \approx \sum_{i=l}^{L} \sigma(\left[\text{MSA}^i(\mZ^{i-1})\right]_{\text{cls}}) \times \ \mT,
\end{equation}
where $L$ denotes the number of transformer layers, $\sigma$ is the projection layer with a fully-connected layer and normalization operation, and MSA stands for multi-head self-attention structure. $\mZ^l$ is input vision tokens for the $l$-th layer, while $[\mZ]_{\text{cls}}$ indicates the class token in sequence $\mZ$.

\subsection{The Composition of Training Data}
\begin{figure}[htbp]
    \centering
    \begin{subfigure}[b]{0.47\textwidth}
        \centering
        \includegraphics[width=\linewidth]{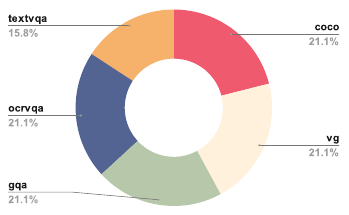}
        \caption{Training data source composition.}
    \end{subfigure}
    \hfill
    \begin{subfigure}[b]{0.47\textwidth}
        \centering
        \includegraphics[width=\linewidth]{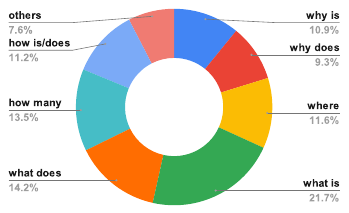}
        \caption{Question type of our training data.}
    \end{subfigure}
    \caption{\textbf{Visual analysis of the composition of training data.} Best viewed in color.}
    \label{appendix:vis}
\end{figure}

Our training source dataset is composed of several components used in LLaVA~\cite{liu2023improvedllava} to ensure that AutoV is trained on a diverse range of image-question pairs. It includes $25$K samples from the COCO train dataset \cite{lin2014microsoft}, $20$K samples from the Visual Genome (VG) dataset \cite{krishna2017visual}, $20$K samples from the GQA dataset \cite{hudson2019gqa}, $20$K samples from the OCRVQA dataset \cite{mishra2019ocr}, and $15$K samples from the TextVQA dataset \cite{singh2019towards}. In total, the source data contains $100$K image-question pairs. Here, we visualize the proportional relationship of the source dataset in Figure \ref{appendix:vis} (a). Furthermore, the question types in our training dataset in visualized in Figure \ref{appendix:vis} (b), and the selected data types are relatively balanced, which enhances the generalizability and robustness of the ranking network in AutoV.

We then generate various visual prompts based on the above source data. Here, we simply use the \texttt{attention prompting} method employed by the API \cite{api} for high-quality generation. To streamline the process, we empirically generate heat maps of attention maps corresponding to CLIP \cite{CLIP} layers $15$, $20$, $22$, and $23$ under its setting on the image as our training visual prompts. 

A specific example (including JSON dictionary and images) can be found in Figure \ref{fig:vis_train}. As a result, one question will correspond to four visual prompts, with a total of AutoV being trained on $40$K visual prompts.

\section{Evaluation Benchmarks}
We conduct comprehensive experiments on \textbf{fifteen} vision-language benchmarks.

\textbf{MMMU.}~\cite{yue2024mmmu} The MMMU is designed to evaluate LVLMs across a wide range of academic disciplines and complex tasks. It covers over $30$ subjects, including history, medicine, physics, and more, requiring multimodal comprehension, fine-grained reasoning, and domain-specific knowledge. Each question is paired with an image and a multiple-choice question, testing the ability to integrate visual understanding with deep textual reasoning.

\textbf{GQA.}~\cite{hudson2019gqa} The GQA benchmark is composed of three parts: scene graphs, questions, and images. The image part contains images, as well as the spatial features of images and the features of all objects in images. The questions in GQA are designed to test the understanding of visual scenes and the ability to reason about different aspects of an image.

\textbf{POPE.}~\cite{li2023evaluating} The POPE benchmark is primarily used to evaluate the degree of Object Hallucination in models. It reformulates hallucination evaluation by requiring the model to answer a series of specific binary questions regarding the presence of objects in images. Accuracy, Recall, Precision, and F1 Score are effectively employed as reliable evaluation metrics to precisely measure the model's hallucination level under three different sampling strategies.

\textbf{TextVQA.}~\cite{singh2019towards} The TextVQA benchmark focuses on the comprehensive integration of diverse text information within images. It meticulously evaluates the model’s text understanding and reasoning abilities through a series of visual question-answering tasks with rich textual information. Models need to not only understand the visual content of the images but also be able to read and reason about the text within the images to answer the questions accurately.

\textbf{MMVet.}~\cite{yu2023mm} The MMVet benchmark is designed based on the insight that the intriguing ability to solve complicated tasks is often achieved by a generalist model that can integrate different core vision-language capabilities. MM-Vet defines 6 core capabilities and examines the 16 integrations of interest derived from the capability combination.

\textbf{LLaVA-In-The-Wild.}~\cite{liu2023improvedllava} The LLaVA-In-The-Wild benchmark is to assess the generalization ability of LVLMs in real-world, unconstrained settings. Unlike curated benchmarks, it features user-collected image-question pairs from diverse sources and domains. This setup presents models with noisy, complex, and unpredictable visual content, offering a realistic measure of their robustness, grounding, and performance in open-ended scenarios.

\textbf{MMStar.}~\cite{chen2024we} MMStar evaluates six vision-language capabilities across 18 fine-grained dimensions using visually dependent questions. It introduces Multi-modal Gain and Multi-modal Leakage metrics to measure visual contribution and data leakage. This benchmark rigorously tests models’ true visual reasoning and generalization abilities.

\textbf{MathVista.}~\cite{lu2023mathvista} MathVista assesses mathematical reasoning grounded in visual contexts such as diagrams and plots. It integrates 28 prior datasets and adds new subsets targeting logical, functional, and scientific figure reasoning. The benchmark challenges models to combine precise visual understanding with compositional calculations and multi-step quantitative reasoning.

\textbf{CVBench.}~\cite{cvbench2025} CVBench measures models’ ability to reason across multiple videos. Its tasks involve object association, event association, and complex multi-video reasoning between distinct video clips. The dataset highlights large performance gaps between humans and models in multi-video reasoning and temporal integration.

\textbf{BLINK.}~\cite{fu2024blink} BLINK targets core perceptual abilities that humans solve instantly but multimodal models struggle with. It reformulates 14 classic vision tasks into 3,800 multiple-choice questions testing fine-grained perception. Results reveal that current models still lack fundamental visual understanding compared to humans.

\textbf{Real-World QA.}~\cite{RealWorldQA} To develop useful real-world AI assistants, it is crucial to advance a model's understanding of the physical world. RealWorldQA is a benchmark just for the design. The dataset consists of anonymized images taken from vehicles, in addition to other real-world images. The initial release of the RealWorldQA consists of over 700 images, with a question and an easily verifiable answer for each image. 

\textbf{MMVP.}~\cite{tong2024eyes} MMVP exposes perceptual blind spots in vision-language models using “CLIP-blind” image pairs. These are pairs of images that contain clear visual differences to humans but are nearly indistinguishable to CLIP-based vision encoders. Leveraging such tricky image pairs, the benchmark pinpoints core weaknesses in visual representation and perception fidelity.

\textbf{COCOCap.}~\cite{lin2014microsoft} Microsoft COCO Image Captioning benchmark, a standard test of a model’s ability to generate natural language descriptions of images. It contains over 330,000 images, each with five human-written captions. Performance reflects a model’s capacity for coherent and contextually accurate visual description.

\textbf{RefCOCO+.}~\cite{kazemzadeh2014referitgame} RefCOCO+ measures models’ ability to ground appearance-based referring expressions in images. Its annotations forbid spatial words, forcing reliance on attributes. This benchmark tests precise visual grounding under complex, attribute-driven conditions.

\textbf{ImageNet-1K.}~\cite{imagenet15russakovsky} ImageNet-1K is a foundational benchmark for large-scale image classification. It contains 1,000 object categories with 1.28M training and 50K validation images. Top-1 and Top-5 accuracy serve as key metrics for evaluating visual recognition performance.

\section{Implementation Details}
\label{details}
(1) \textbf{Data generation stage}: LLaVA-v1.5 7B is consistently adopted as the reference LVLM for generating loss due to resource savings. In total, 40K samples require loss annotation. With standard single A100-40G GPU inference, the entire process can be completed within half a day, making the data generation stage computationally manageable. 

\noindent(2) \textbf{AutoV training stage}: The LLaVA series is trained using DeepSpeed \cite{rasley2020deepspeed} ZeRO2 with $8$ A100-40G GPUs within $40$ epochs. The batch size (with accumulation) is set to $64$, and the learning rate is $1e$-$3$. For convenience, Qwen2.5-VL, InternVL2, Gemini-1.5-Pro, GPT-4o, and Qwen-VL-Max directly adopt the retrieval strategy in AutoV pre-trained from LLaVA-OneVision.

\section{Reliability of Language-modeling Loss}
We mitigate language priors using a low-variance criterion, where \textbf{small loss variation across images indicates weak visual dependence}. For instance, we conduct a query-only evaluation on the training set, where accuracy drops from $32.7\%$ before filtering to $13.2\%$ after filtering, indirectly indicating stronger visual grounding in the filtered data. Besides, we adopt the GPT-4o to judge language-prior cases, where the proportion decreases from $27.5\%$ to $8.1\%$ after filtering. Therefore, these results provide stronger empirical support.

\begin{table}[t]
\small
\centering
\setlength{\tabcolsep}{1.7pt}
\renewcommand{\arraystretch}{1.3}
\caption{\textbf{AutoV with various teacher model scales.}}
\label{tab:teacher_scale}
\begin{tabular}{lcc|lcc}
    \toprule
    \textbf{Models} & \textbf{VizWiz} & \textbf{LLaVA}$^{\text{Wild}}$ 
    & \textbf{Models} & \textbf{VizWiz} & \textbf{LLaVA}$^{\text{Wild}}$ \\
    \midrule

    \raisebox{-0.25\height}{\includegraphics[height=1em]{fig/chatgpt.png}}\hspace{0.2em} GPT-4o & 52.8 & 68.0 
    & \raisebox{-0.25\height}{\includegraphics[height=1em]{fig/qwen.png}}\hspace{0.2em} Qwen-Max & 64.6 & 85.8 \\

    \rowcolor{mygray}
    \multicolumn{6}{c}{Policy from LLaVA-OneVision (\textbf{0.5B})} \\
    + AutoV & 56.9\gain{4.1} & 69.7\gain{1.7} 
            & + AutoV & 68.9\gain{4.3} & 87.0\gain{1.2} \\

    \rowcolor{mygray}
    \multicolumn{6}{c}{Policy from LLaVA-OneVision-1.5 (\textbf{4B})} \\
    + AutoV & 61.4\gain{8.6} & 70.3\gain{2.3} 
            & + AutoV & 71.1\gain{6.5} & 88.1\gain{2.3} \\

    \bottomrule
\end{tabular}
\end{table}

\section{Extended Evaluations on Additional Benchmarks}
\label{sec:extended_eval}
To further verify the generalization capability of AutoV, we extend our evaluation to a broader range of seven vision-language benchmarks. These include CVBench, BLINK, RealWorldQA, MMVP, COCOCap, RefCOCO+, and ImageNet-1K, covering diverse tasks such as fine-grained visual perception, multi-modal reasoning, open-ended captioning, and image classification. 
As shown in Table~\ref{tab:appendix_extra}, we conduct these extended experiments using two state-of-the-art 7B-parameter vision-language models as backbones: LLaVA-OneVision 7B~\cite{li2024llava} and Qwen2.5-VL 7B~\cite{bai2025qwen2}. 
The results consistently demonstrate that AutoV enhances multimodal understanding and reasoning across all evaluated domains. Specifically, for LLaVA-OneVision 7B, AutoV achieves average improvements ranging from $+2.6\%$ to $+3.5\%$ over the base model. A similar consistent performance boost is observed when employing Qwen2.5-VL 7B, where AutoV improves performance across all tasks (\textit{e.g.}, $+3.0\%$ on MMVP and $+2.6\%$ on RefCOCO+). This outstanding cross-task generalization highlights the robust capability of AutoV in multi-modal retrieval and representation alignment.

\begin{table*}[t]
    \centering
    \setlength{\tabcolsep}{1.3pt}
    \renewcommand{\arraystretch}{1.2}
    \footnotesize
    \caption{
    \textbf{Extended evaluation of AutoV on additional benchmarks.}}
    \vspace{-2mm}
    \label{tab:appendix_extra}
\resizebox{.97\linewidth}{!}{
\begin{tabular}{l | c c c c | c c c}
\shline
\textbf{Method} & \textbf{CVBench} & \textbf{BLINK} & \textbf{RealWorld} & \textbf{MMVP} & \textbf{COCOCap} & \textbf{RefCOCO+} & \textbf{ImageNet-1K} \\
\hline

\multicolumn{8}{c}{\raisebox{-0.2\height}{\includegraphics[height=1.em]{fig/llava.png}}\hspace{0.2em} \textbf{(LLaVA-OneVision 7B)}}\\
Base & 79.6 & 48.2 & 66.0 & 60.7 & 61.1 & 68.8 & 92.3 \\
API  & 80.7 \gain{1.1} & 49.4 \gain{1.2} & 66.4 \gain{0.4} & 61.5 \gain{0.8} & 61.9 \gain{0.8} & 69.4 \gain{0.6} & 92.6 \gain{0.3} \\
\textbf{AutoV} & \textbf{83.0} \gain{\textbf{3.4}} & \textbf{51.6} \gain{\textbf{3.4}} & \textbf{68.9} \gain{2.9} & \textbf{64.2} \gain{\textbf{3.5}} & \textbf{63.7} \gain{2.6} & \textbf{71.8} \gain{\textbf{3.0}} & \textbf{95.0} \gain{2.7} \\
\hline

\multicolumn{8}{c}{\raisebox{-0.2\height}{\includegraphics[height=1.4em]{fig/qwen.png}}\hspace{0.2em} \textbf{(Qwen2.5-VL 7B)}}\\
Base & 80.6 & 56.4 & 68.5 & 61.3 & 40.7 & 72.6 & 94.7 \\
API  & 81.1 \gain{0.5} & 57.3 \gain{0.9} & 69.2 \gain{0.7} & 61.7 \gain{0.4} & 41.4 \gain{0.7} & 73.5 \gain{0.9} & 95.4 \gain{0.7} \\
\textbf{AutoV} & \textbf{83.0} \gain{2.4} & \textbf{59.0} \gain{2.6} & \textbf{71.1} \gain{2.6} & \textbf{64.3} \gain{\textbf{3.0}} & \textbf{43.3} \gain{2.6} & \textbf{75.2} \gain{2.6} & \textbf{97.1} \gain{2.4} \\
\shline
\end{tabular}
}
\end{table*}

\section{Analysis of the Teacher (Strategy)  Model Scale}
We study the effect of teacher scale using smaller LLaVA-OneVision series models (0.5B and 4B). As shown in Table \ref{tab:teacher_scale}, AutoV consistently improves performance on GPT-4o, \eg, $+4.1\sim+8.6$ on VizWiz and $+1.7\sim+2.3$ on LLaVA$^{\text{Wild}}$, even with the smallest $0.5$B teacher. Therefore, AutoV is robust to teacher scale, while larger teachers provide slightly higher gains, enabling flexible efficiency–performance trade-offs for users.

\begin{table}[t]
\setlength\tabcolsep{6mm}
\small
\centering
\caption{\textbf{Runs on MMMU task with various seeds.}}
\label{tab:runs}
\begin{tabular}{c|ccc|c}
\toprule
\textbf{Run} & \textbf{Seed} & \textbf{AutoV} & \textbf{Baseline} & \textbf{Diff.} \\
\midrule
1   & 10   & 38.7 & 36.3 & +2.4 \\
2   & 20   & 38.7 & 36.3 & +2.4 \\
3   & 30   & 38.8 & 36.2 & +2.6 \\
4   & 40   & 38.7 & 36.2 & +2.5 \\
5   & 50   & 38.8 & 36.4 & +2.4 \\
\bottomrule
\end{tabular}
\end{table}

\section{Statistical Significance Analysis}
\label{sec:ss}
To further verify the robustness of the improvement in the main experiments, we conduct significance testing across multiple random seeds on representative benchmarks. Specifically, we present the detailed analysis process for MMMU as an example, while the same manner applies to the remaining benchmarks. The results are shown in Table \ref{tab:runs}, and the analysis is presented below:

\begin{center}
\begin{itemize}
    \item \textbf{Mean} $\Delta$: +2.06\%
    \item \textbf{Std Dev}: $\pm 0.083\%$
    \item $p$-\textbf{value}: $< 0.05$
\end{itemize}
\end{center}

The above analysis provides \textbf{solid statistical evidence} that AutoV delivers consistent and reproducible improvements across runs, highlighting its robustness and reliability in enhancing vision-language model performance.

\section{More Visualization Results}
\label{morevis}
\subsection{Cases of Failure}
\begin{figure*}[t]
    \centering
    \includegraphics[width=0.98\textwidth]{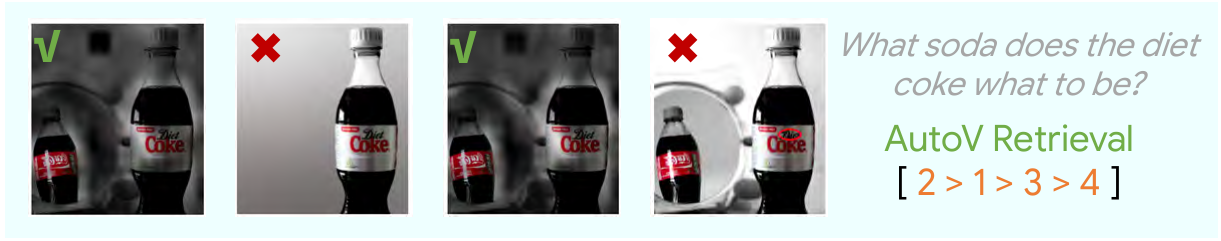}
    \caption{\textbf{Visualization of a failure case.}}
    \label{fig:failure}
\end{figure*}
As shown in Fig.~\ref{fig:failure}, due to the abstract nature of the query, AutoV is misled by semantic relevance and incorrectly selects VP\_2, while overlooking the visual cue reflected in the mirror.

\subsection{Cases of Training Data}
\begin{figure*}[htbp]
    \centering
    \includegraphics[width=0.98\textwidth]{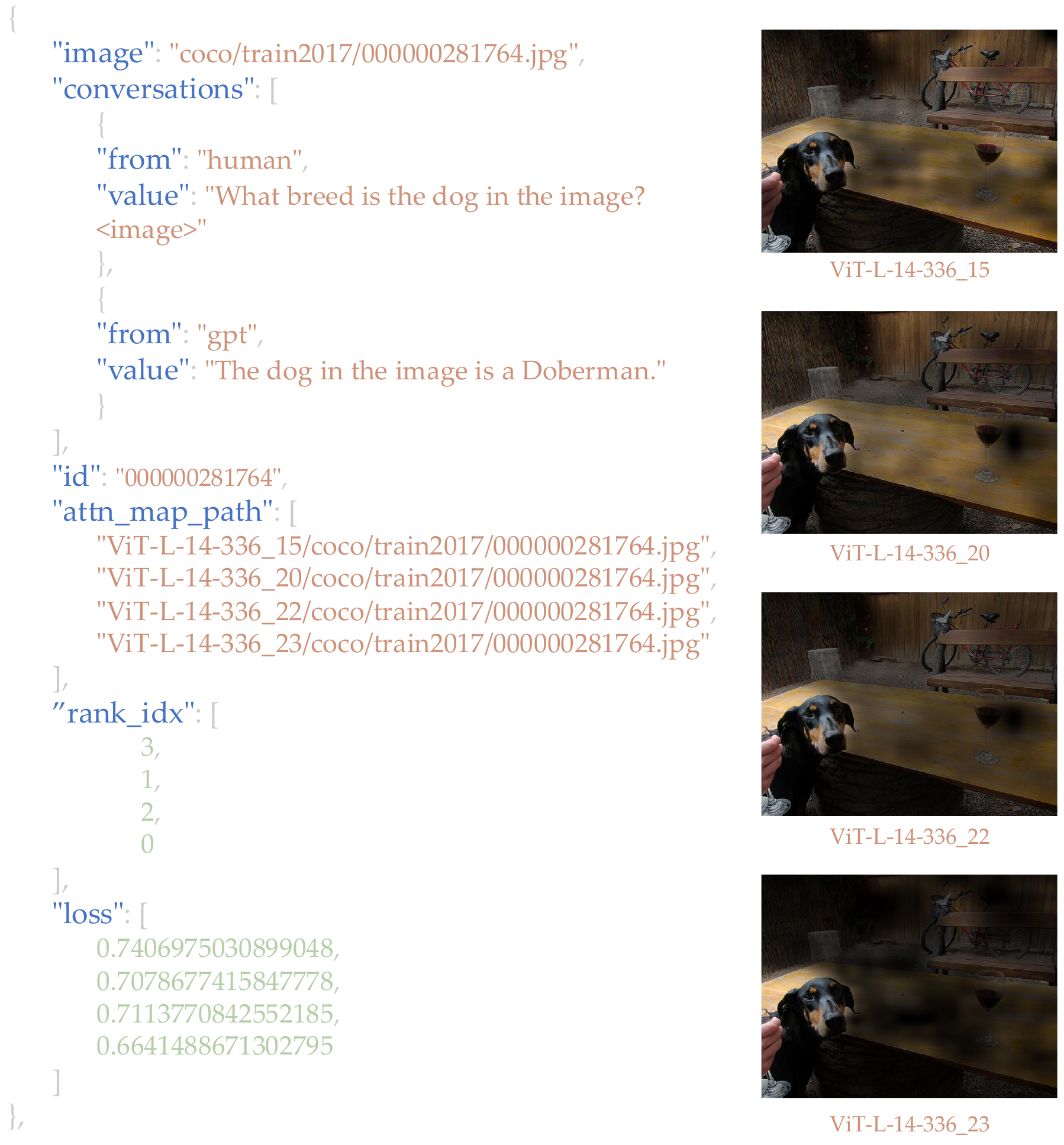}
    \caption{\textbf{Visualization of a training sample in AutoV.} Best viewed in color.}
    \label{fig:vis_train}
\end{figure*}

Figure~\ref{fig:vis_train} shows one training example in our dataset, including its JSON structure and the corresponding attention maps. The JSON entry stores the image path, human–GPT conversation, different visual prompt paths from the API, as well as the predicted ranking and loss values. The right side visualizes these visual prompts, illustrating how different prompts emphasize varying regions of the same image.

\end{document}